\documentclass{article}

\usepackage[preprint]{cs224r_2025}
\usepackage[utf8]{inputenc} 
\usepackage[T1]{fontenc}    
\usepackage{hyperref}       
\usepackage{url}            
\usepackage{booktabs}       
\usepackage{amsfonts}       
\usepackage{nicefrac}       
\usepackage{microtype}      
\usepackage{xcolor}         
\usepackage{amsmath}
\usepackage{graphicx}       
\usepackage{lipsum}         

\title{Reinforcement Learning Fine-Tuning for Instruction Following and Math Reasoning}

\author{%
  Yifu Han \\
  Department of Energy Science \& Engineering\\
  Stanford University\\
  \texttt{yifu@stanford.edu} \\
  \And
  Geo Zhang \\
  Department of Energy Science \& Engineering\\
  Stanford University\\
  \texttt{gmzhang@stanford.edu} \\
}

\begin{document}

\maketitle

\begin{abstract}
This study investigates the effectiveness of reinforcement learning (RL) fine-tuning techniques on a compact language model (Qwen2.5-0.5B Base) for two challenging tasks: instruction following and mathematical reasoning. We compare supervised fine-tuning (SFT), Direct Preference Optimization (DPO) using preference-labeled data, and Reinforce Leave-One-Out (RLOO) with reward models. Our experiments show that RLOO with DeBERTa reward modeling achieves the best alignment, while DPO provides strong and consistent results. For math reasoing tasks, synthetic data augmentation and best-of-N sampling with an external verifier significantly improve accuracy, showing the potential of combining fine-tuning with inference-time tools. This study highlights key trade-offs and practical strategies for training lightweight, task-aligned small-scale language models.
\end{abstract}

\section{Introduction}

Generative language models have demonstrated impressive capabilities in various natural language processing (NLP) tasks, ranging from open-domain question answering to logical inference and symbolic mathematics. However, achieving high performance across diverse domains still needs to be explored. In addition, different fine-tuning techniques including RL should be compared.

In this study, we explore the implementation of RL algorithms for fine-tuning language models (LMs) to improve their performance on two tasks, instruction following and mathematical reasoning. Instruction following~\citep{cui2023ultrafeedback} task is designed to test the model ability to generate high quality responses using natural language prompts, especially if the instruction is unseen in the training set and of different domain and difficulty levels. For mathematical reasoning, Countdown game is considered, which is a task to combine input numbers with arithmetic operations to get a target number~\citep{gandhi2024countdown}. This task will evaluate the model ability to in solving math problems. We focus on training a relatively small language model, Qwen2.5-0.5B Base~\citep{hui2024qwen2}, to assess the performance on various fine-tuning techniques. Our goal is to better understand the capabilities and limitations of lightweight LMs in human-aligned learning settings.

We apply three different approaches: supervised fine-tuning (SFT), Direct Preference Optimization (DPO)~\citep{rafailov2023dpo}, and Reinforce Leave-One-Out (RLOO)~\citep{ahmadian2024reinforce}. These methods are used to fine-tune language model on instruction-following task, where the objective is to produce responses that align well with human preference. In the case of RLOO, we evaluate different reward models, including DeBERTa~\citep{he2020deberta}, DistilBERT~\citep{sanh2019distilbert}, and Siamese DistilBERT~\citep{reimers2019sBERT}. The DistilBERT and Siamese DistilBERT reward models are trained to evaluate alignment quality, and their impact on final policy performance is compared.

To further improve the mathematical reasoning capabilities of the fine-tuned Qwen model, we construct a high-quality synthetic dataset consisting of 1600 examples based on the Countdown dataset. This dataset considers multi-step numerical reasoning and arithmetic composition. We apply GPT-4o~\citep{hurst2024gpt4o} to assist with both problem generation and solution verification, ensuring the training data quality. In addition, we experiment with a best-of-N sampling strategy~\citep{chow2024inference}, guided by an external verifier, to improve the reliability and correctness of model predictions. This approach significantly improves model performance on the math reasoning task, suggesting that combining fine-tuned policies with external tools can generate more accurate results.

Overall, this work contributes practical insights into the trade-offs between different RL-based fine-tuning algorithms for language models and demonstrates effective strategies for improving performance in challenging downstream tasks such as instruction following and mathematical reasoning.

\section{Related Work}

Using external tools is an important approach to improve the performance of LLMs. \citet{jin2025Search-r1} proposed Search-R1, an RL framework that combines search engine interaction with proximal policy optimization (PPO) and group relative policy optimization (GRPO) training, which generates better results compared with the traditional retrieval-augmented generation methods. \citet{gehring2024rlef} introduced an executive feedback workflow, improving code generation capability in external test cases. \citet{schick2023toolformer} showed that LLMs can learn to use various tools correctly through fine-tuning, learning when and how to call the appropriate APIs for each task.

The use of best-of-N sampling guided by an external verifier can be used for improving the reliability and correctness of language model outputs. This strategy involves generating multiple candidate responses from the model and selecting the best based on external evaluation criteria~\cite{chow2024inference}. \citet{zhou2023least} proposed the least-to-most prompting and demonstrated that combining multiple outputs with verification significantly improve reasoning performance. Similarly, \citet{madaan2023selfrefine} introduced self-refine, where a verifier model iteratively improves responses by selecting the best between generated candidates. \citet{zhang2023language} further explored this direction by training verifier models to score and rerank candidate outputs in a language model cascades framework, which generates more factual and consistent generations. These methods allow for more accurate and robust outputs, which is important in mathematical reasoning and code generation. Our work applies best-of-N sampling with an external verifier to improve accuracy in arithmetic composition tasks.

\section{Method}

DPO is a reinforcement learning method for aligning language models with human preferences by directly optimizing a loss that favors preferred responses over less preferred ones. Given a dataset of preference pairs \((x, y^+, y^-)\), where \(x\) is the prompt, \(y^+\) is the preferred response, and \(y^-\) is the less preferred one, DPO minimizes a contrastive loss derived from the reward difference between responses. Assuming the reward function is the log-ratio between the fine-tuned policy \(\pi_\theta\) and a reference policy \(\pi_{\text{ref}}\), the objective becomes:
\[
\mathcal{L}_{\text{DPO}}(\theta) = -\log \left[ \frac{\exp\left(\beta \cdot \log \frac{\pi_\theta(y^+ \mid x)}{\pi_{\text{ref}}(y^+ \mid x)}\right)}{\exp\left(\beta \cdot \log \frac{\pi_\theta(y^+ \mid x)}{\pi_{\text{ref}}(y^+ \mid x)}\right) + \exp\left(\beta \cdot \log \frac{\pi_\theta(y^- \mid x)}{\pi_{\text{ref}}(y^- \mid x)}\right)} \right]
\]
where \(\beta\) is a temperature parameter that controls the sharpness of preference weighting. This DPO implementation allows stable and efficient preference optimization without reward modeling.

\begin{figure}[h!]
  \centering
  \includegraphics[width=0.8\linewidth]{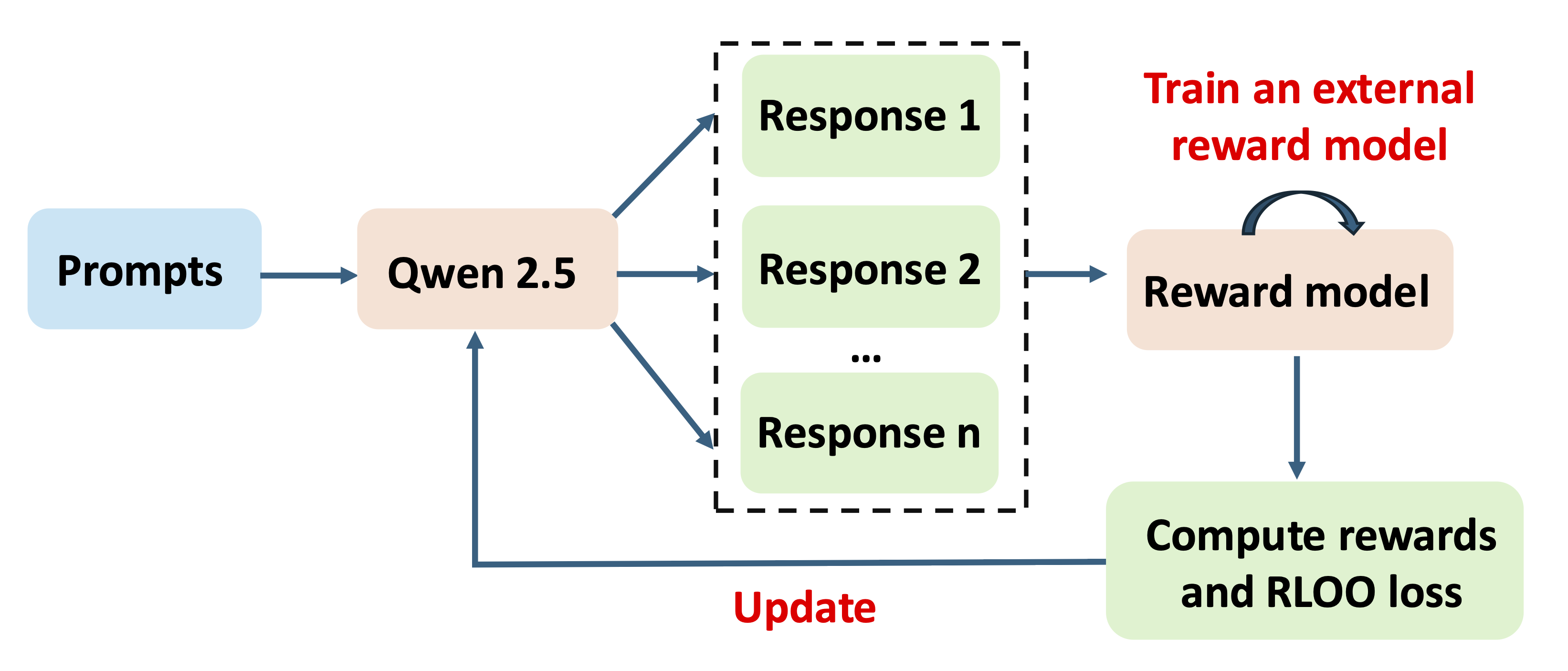}
  \caption{Overview of RLOO method with an external reward model trained and uilized for scorer. }
  \label{fig:RLOO}
\end{figure}

RLOO is a policy gradient based algorithm to fine-tune language models by reducing the variance of gradient estimates using a leave-one-out baseline. Given a set of \(N\) sampled responses \(\{y_1, \ldots, y_N\}\) to a prompt \(x\), each response \(y_i\) is assigned a scalar reward \(r_i\), computed using a separate reward model or preference function, as shown in Fig. \ref{fig:RLOO}. The gradient estimate for policy \(\pi_\theta\) is computed as:

\[
\nabla_\theta \mathcal{L}_{\text{RLOO}}(\theta) = -\frac{1}{N} \sum_{i=1}^N \left( r_i - \frac{1}{N-1} \sum_{j \ne i} r_j \right) \nabla_\theta \log \pi_\theta(y_i \mid x)
\]
where \(\frac{1}{N-1} \sum_{j \ne i} r_j\) is the leave-one-out baseline, effectively reducing variance in the gradient estimate. This enables stable and sample efficient fine-tuning of language models with explicit preference to the high-reward responses.

In RLOO, a critical component is the reward model, which assigns scores to generated responses. The reward model can be based on human feedback, heuristic algorithms, or learned models. In this work, we use pretrained transformer-based models, specifically variants of BERT (Bidirectional Encoder Representations from Transformers)~\citep{devlin2019bert}, which is effective in a wide range of NLP tasks. To explore the trade-offs between performance and efficiency, we applied two another BERT variants, (1) DeBERTa (Decoding-enhanced BERT with disentangled attention)~\citep{he2020deberta}, which has strong representation capabilities and improved generalization due to the enhanced attention mechanism and disentangled positional encoding; (2) DistilBERT~\citep{sanh2019distilbert}, due to its computational efficiency. These models are adapted by adding a regression head to output scalar reward scores and fine-tuned on the offline UltraFeedback dataset~\cite{cui2023ultrafeedback}. In addition to the standard single-input reward model, we also tested a Siamese BERT structure~\citep{reimers2019sBERT} for direct preference modeling. Both the preferred and dispreferred responses (conditioned on the same prompt) are passed independently through a shared BERT encoder to produce two scalar scores and then the model is trained using the Bradley-Terry objective, which is expressed as

\[
\mathcal{L}_{\text{BT}}(\phi) = \max_{\phi} \; \mathbb{E}_{(x, y^+, y^-) \sim \mathcal{D}_{\text{pref}}} \left[ \log \sigma\left( r_{\phi}(x, y^+) - r_{\phi}(x, y^-) \right) \right]
\]

\begin{figure}[h!]
  \centering
  \includegraphics[width=0.8\linewidth]{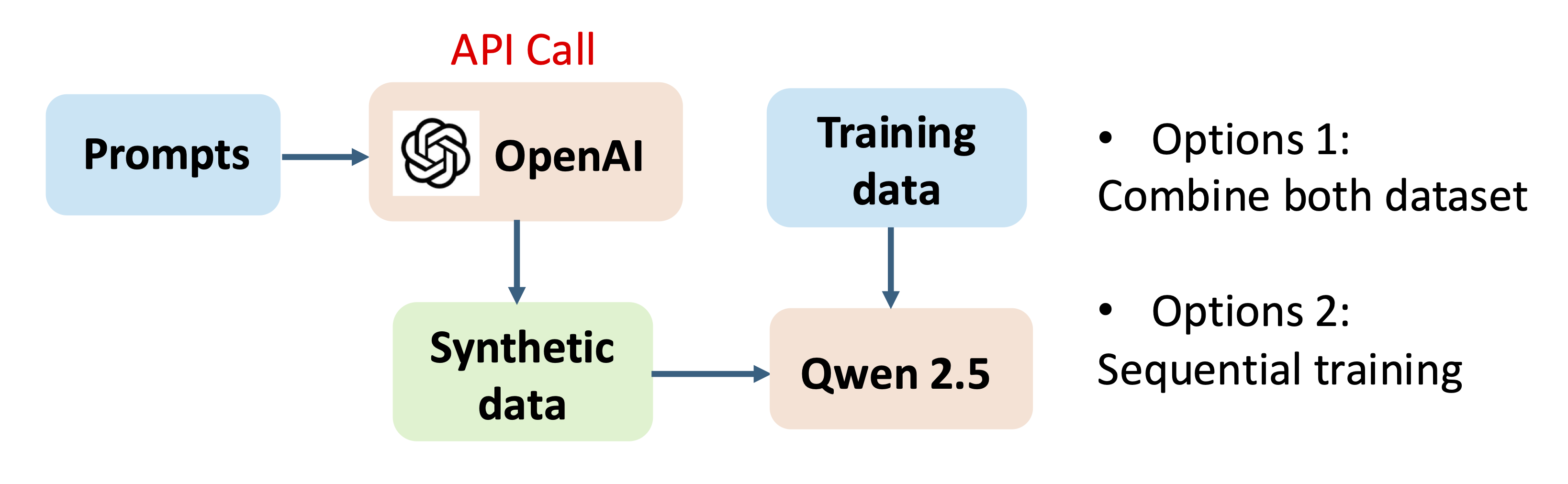}
  \caption{Generation of synthetic data using ChatGPT and utilization in model training. }
  \label{fig:Synthetic}
\end{figure}

On the math reasoning task, due to the relative small size of the dataset, we explored using synthetic data to augment the fine-tuning procedure. The synthetic data is generated through input prompts into GPT-4o~\citep{hurst2024gpt4o} as shown the process in Fig. \ref{fig:Synthetic}.

\begin{figure}[h!]
  \centering
  \includegraphics[width=0.8\linewidth]{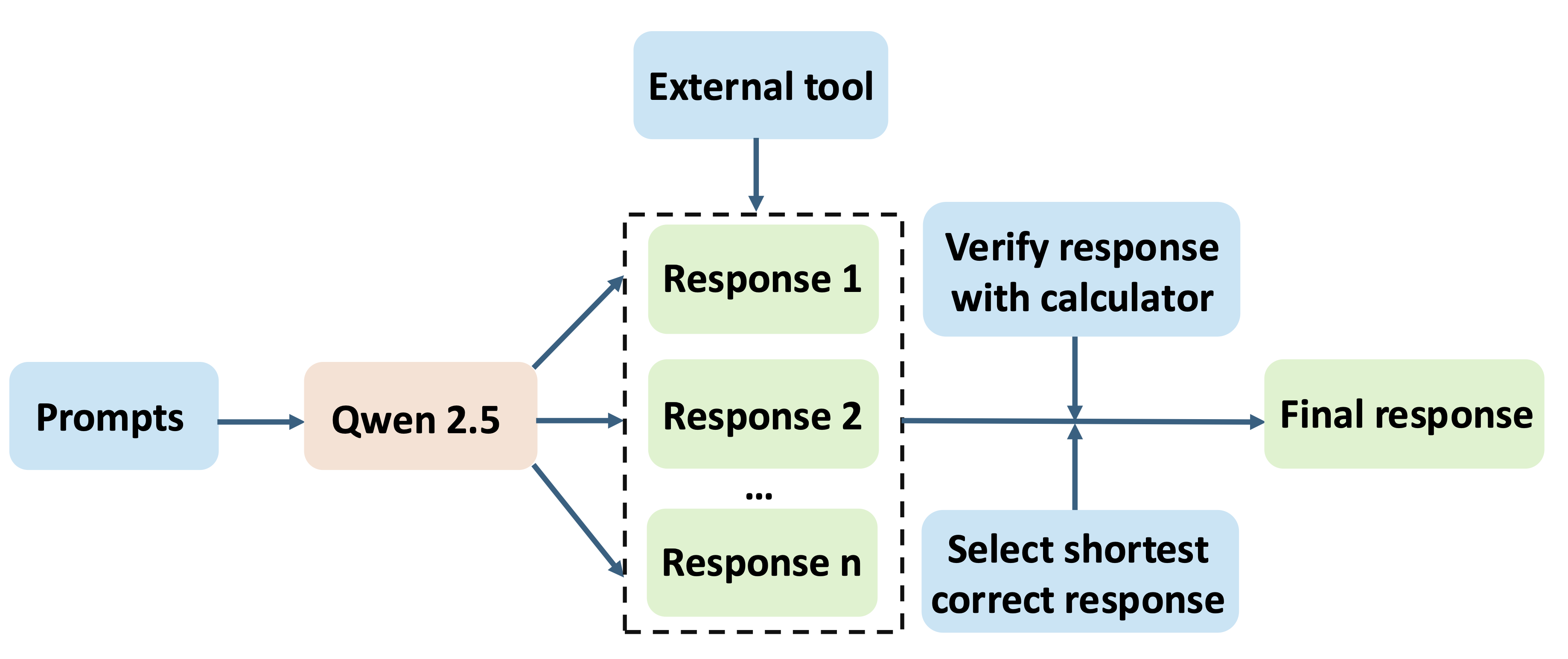}
  \caption{Use best-of-N sampling strategy and external tool as critics.}
  \label{fig:external}
\end{figure}

To improve the capabilities, we implemented a best-of-N sampling strategy~\citep{chow2024inference} and incorporated an external tool as critics to select the most appropriate response, as illustrated in Fig.~\ref{fig:external}.

\section{Experimental Setup}

We have performed two supervised fine-tuning runs for the Qwen 2.5-0.5B baseline model on a single NVIDIA A100 GPU: a LoRA run for instruction following task on the SmolTalk corpus (learning-rate 5e-6, 3 epochs, $\approx$11 h) and a full-parameter run for math reasoning on the Countdown Warmstart data (same learning rate, 5 epochs, $\approx$10 min). Evaluation procedure used vLLM for generation and was performed on 200 randomly selected SmolTalk test samples, and 1000 randomly sampled Countdown-Tasks-3to4 problems. These two cases are then scored with the Llama 3.1 Nemotron-70B reward model and an exact-match countdown.py script, respectively.

\begin{figure}
    \centering
    \includegraphics[width=0.95\linewidth]{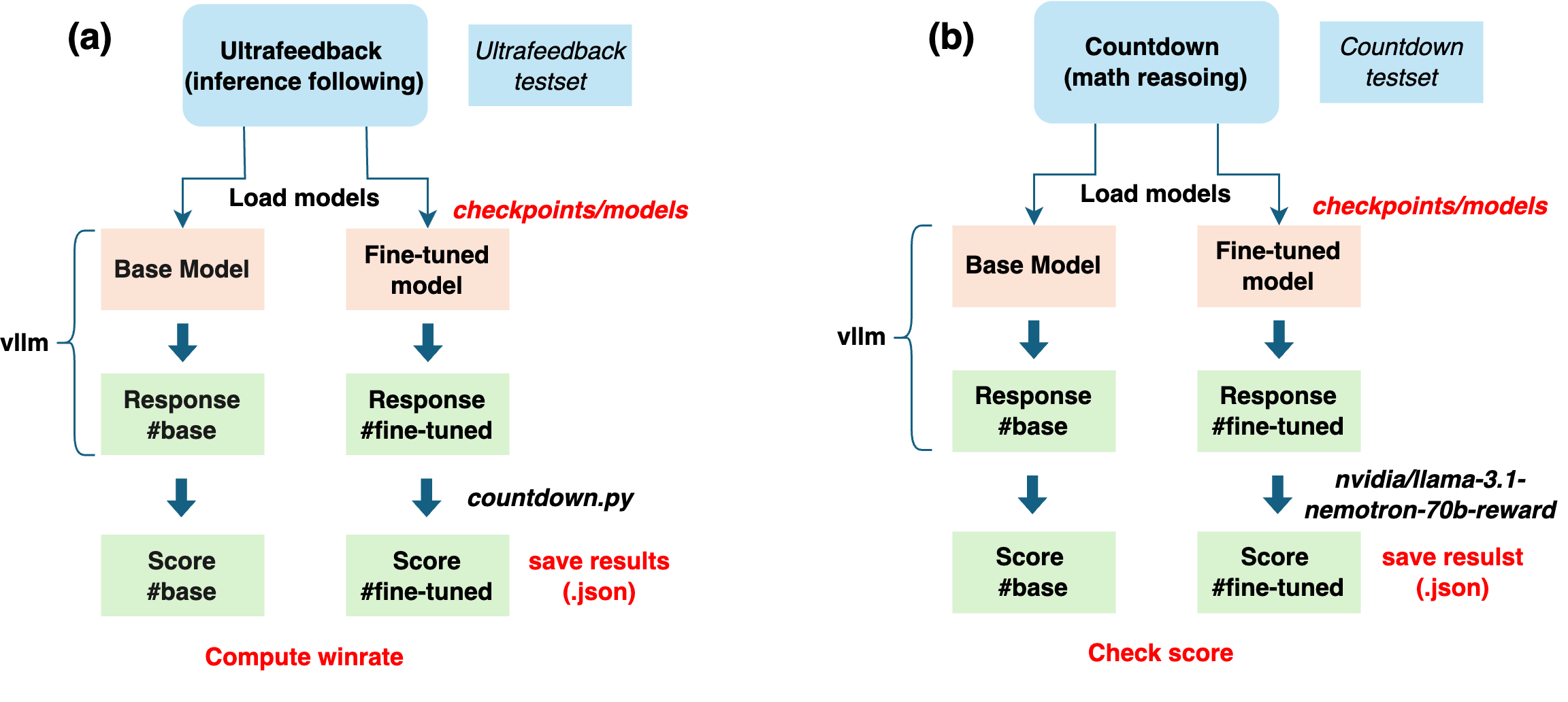}
    \caption{The evaluation process of two tasks }
    \label{fig:enter-label}
\end{figure}

\section{Results}

\subsection{Quantitative Evaluation}

Table.\ref{rl} compares the performance of different methods including SFT, DPO, and RLOO using both full and LoRA fine-tuning setups, as well as different reward functions for RLOO. The results show the winning-rate evaluation metric, where DPO achieves higher winning-rate than SFT in both full (0.605 compared with 0.495) and LoRA (0.665 compared with 0.575) settings. Between RLOO variants, the model trained with reward model of DeBERTa achieves the highest score (0.695), which is better than both SFT and DPO. DPO consistently generate better results compared with SFT. We also find that by using the LoRA adapters with the rank of 8,we achieved comparable performance than full parameter fine-tuning while reducing GPU memory and training time more than 50\%.

\begin{table}[h!]
\centering
\caption{Comparison of models using different training methods and reward functions.}
\begin{tabular}{p{1.6cm} p{2cm} p{2cm} p{1.8cm} p{1.8cm} p{1.8cm}}
\toprule
& SFT \newline (full/LoRa) & DPO \newline (full/LoRa) & RLOO \newline ($\text{reward}^1$) & RLOO \newline ($\text{reward}^2$) & RLOO \newline ($\text{reward}^3$) \\
\midrule
Score & 0.495 / 0.575 & 0.605 / 0.665 & 0.695 & 0.530 & 0.535 \\
\bottomrule
\end{tabular}
\label{rl}
\end{table}

\begin{table}[h!]
\centering
\caption{Comparison of SFT (full parameters fine-tuning) with synthetic data and using external tool.}
\begin{tabular}{lccc}
\toprule
& SFT (full) & Synthetic Data (SFT full) & External Tool (SFT full) \\
\midrule
Score & 0.37 & 0.3835 & 0.811 \\
\bottomrule
\end{tabular}
\label{math}
\end{table}

Table.\ref{math} presents a comparison of performance scores on the math reasoning task. The baseline model, trained with SFT data, achieves a score of 0.37. When synthetic data is incorporated into the same SFT framework, performance improves slightly to 0.3835, indicating a modest benefit from augmenting the training set. However, the most significant improvement comes from integrating an external tool at inference time, which boosts the score substantially to 0.811. This highlights the effectiveness of using an external verifier or reasoning tool to select or validate generated answers, showing that test time inference can achieve significantly improve performance and accuracy.

\begin{figure}[h!]
    \centering
    \includegraphics[width=0.85\linewidth]{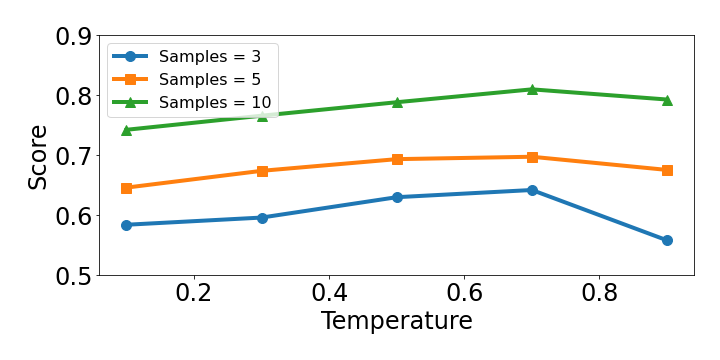}
    \caption{Scores achieved for the math reasoning task by using the external verifier for different numbers of best-of-N sampling and temperature. }
    \label{fig:n-sampling}
\end{figure}

Fig.\ref{fig:n-sampling} illustrates the impact of temperature and the number of samples used in best-of-N sampling on the performance in the math reasoning task, as evaluated by an external verifier. In this experiment, we evaluate three sample sizes: 3, 5, and 10. The results show that increasing the number of samples consistently improves performance over all temperature settings, with the best scores achieved when using 10 samples. As temperature increases, the scores initially improve due to greater sample diversity, but tend to decline at higher temperatures, particularly when using fewer samples. The highest score is obtained with 10 samples at a temperature of about 0.7. This suggests that using moderate temperature with a larger sample allows the external verifier to more effectively select accurate solutions during the test-time.

\subsection{Qualitative Analysis}

To better understand model behavior across tasks, we conducted qualitative analysis of representative model output from each fine-tuning model.

For the instruction-following task, output from SFT model tend to be generic or overly verbose, occasionally missing the intent of the prompt. While the model fine-tuned by DPO often generates more precise and human-aligned answers, due to its preference-based optimization. For example, when asked to summarize a technical concept, DPO responses were more concise and accurate than SFT. The RLOO model, using DeBERTa reward model, generates outputs with more consistent tone and relevance, although it occasionally produces repetitive phrasing due to reward overfitting.

In the math reasoning task, the base SFT model frequently fails to complete all computation steps or makes arithmetic mistakes. The inclusion of GPT-4o-generated synthetic examples improves multi-step reasoning slightly. However, the most significant improvement arises from applying best-of-N sampling with an external verifier. For example, when asked to reach a target number using a set of integers, the verifier consistently selected correct outputs from between diverse candidates.

These results indicate that while fine-tuning improves alignment and task specificity, inference-time strategies, such as verifier-guided sampling provide a powerful, complementary mechanism for ensuring high-quality output, especially for precision-sensitive domains like mathematical reasoning.

\section{Discussion}

For the RLOO algorithm, two factors are essential for its performance: the number of responses generated per query and the quality of the reward function. The number of sampled responses evaluated during training directly influences the model capacity for exploration. Generating more responses per prompt allows the learning algorithm to better explore the policy space, but this comes at the cost of increased computational resources. Second, the reward function is important for guiding the model learning for RLOO. Reward models trained on high-quality, and task specific preference data can better capture differences in reasoning quality, leading to more consistent and reliable improvements for RLOO during the test-time. The use of an external verifier for best-of-N sampling can significantly improve model performance in mathematical reasoning. A larger number of samples and a relatively higher temperature (e.g., 0.7) tend to generate more accurate responses.

\section{Conclusion}

This study explores the application of reinforcement learning fine-tuning techniques to a small-scale language model (Qwen2.5-0.5B Base) on two distinct tasks: instruction following and math reasoning. Through comparative analysis of SFT, DPO, and RLOO, we show that preference-based optimization methods, particularly DPO and RLOO with appropriate reward models, can significantly improve alignment with human feedback, even for lightweight language models.

Our experiments highlight that DeBERTa-based reward models are especially effective in the RLOO framework, while DPO achieves consistently strong performance with a simpler optimization setup. For the math reasoning task, we show that synthetic data generation using GPT-4o marginally improve performance, but the most substantial improvement comes from best-of-N sampling guided by external verifiers. This approach enables higher accuracy without additional training cost.

Overall, our findings suggest that small LMs can achieve competitive results when combined with effective fine-tuning algorithms, high-quality reward modeling, and inference-time verification. This provides a practical path for deploying efficient and aligned language models in low computation resource settings. Future work may explore extending this framework to additional domains such as code generation, symbolic logic, and multimodal tasks.

\bibliographystyle{ACM-Reference-Format}
\bibliography{reference}

\appendix

\section{Implementation Details}
The code is available at {https://github.com/Yifu93/LLM-Reinforcement-Learning}.

\end{document}